\documentclass[conference]{IEEEtran}
\IEEEoverridecommandlockouts

\usepackage{cite}
\usepackage{amsmath,amssymb,amsfonts}
\usepackage{algorithmic}
\usepackage{graphicx}
\usepackage{textcomp}
\usepackage{xcolor}
\usepackage[utf8]{inputenc} 
\usepackage{tabularx} 
\usepackage{cleveref} 
\usepackage{booktabs} 
\usepackage{wrapfig}

\def\BibTeX{{\rm B\kern-.05em{\sc i\kern-.025em b}\kern-.08em
    T\kern-.1667em\lower.7ex\hbox{E}\kern-.125emX}}
\begin{document}

\title{LawDNet: Enhanced Audio-Driven Lip Synthesis via Local Affine Warping Deformation\\
}


\author{
Junli Deng\textsuperscript{1}, Yihao Luo\textsuperscript{2}, Xueting Yang\textsuperscript{3}, Siyou Li\textsuperscript{4}, Wei Wang\textsuperscript{5}, Jinyang Guo\textsuperscript{6}, Ping Shi\textsuperscript{1} \\
\textsuperscript{1}Communication University of China, Beijing, China \\
\textsuperscript{2}Imperial College London, London, UK \\
\textsuperscript{3}Hong Kong University, Hong Kong, China \\
\textsuperscript{4}Queen Mary University of London, London, UK \\
\textsuperscript{5}Beijing University of Posts and Telecommunications, Beijing, China \\
\textsuperscript{6}Beihang University, Beijing, China \\
}

\maketitle

\vspace{-1.2 em}

\begin{abstract}
In the domain of photorealistic talking head generation, the fidelity of audio-driven lip motion synthesis is essential for realistic virtual interactions. Existing methods face two key challenges: a lack of vivacity due to limited diversity in generated lip poses and noticeable anamorphose motions caused by poor temporal coherence. To address these issues, we propose LawDNet, a novel deep-learning architecture enhancing lip synthesis through a Local Affine Warping Deformation mechanism. This mechanism models the intricate lip movements in response to the audio input by controllable non-linear warping fields. These fields consist of local affine transformations focused on abstract keypoints within deep feature maps, offering a novel universal paradigm for feature warping in networks. Additionally, LawDNet incorporates a dual-stream discriminator for improved frame-to-frame continuity and employs face normalization techniques to handle pose and scene variations. Extensive evaluations demonstrate LawDNet's superior robustness and lip movement dynamism performance compared to previous methods. 
\end{abstract}

\begin{IEEEkeywords}
Lip Synthesis, Local Affine Transformations.
\end{IEEEkeywords}

\section{Introduction}
Lip synthesis is important in applications such as digital human creation~\cite{im2023case}, audio dubbing~\cite{brannon2023dubbing}, and entertainment~\cite{zhen2023human, sha2023deep, gowda2023pixels,kadam2021survey,meng2024comprehensive}. Previous methods can be categorized into direct generation and warping-based methods. Direct generation approaches~\cite{gupta2023towards,DBLP:journals/corr/abs-2008-10010, gupta2023towards,talkingicassp2024} rely on up-sampling layers to synthesize pixels from latent embeddings, combining audio information with identity features, but often struggle with frame continuity and preserving unique lip characteristics.

Warping-based methods fall into two categories. The first category uses motion field prediction networks, such as optical flow~\cite{baker2011database}, to generate deformation fields from audio inputs~\cite{yin2022styleheat,zhang2021flow, wang2021audio2head,song2024adaptive}. These methods manipulate pixels on shallower feature maps, effectively preserving individual textures but sometimes causing background distortions. The second category employs specific spatial transform operators~\cite{zhang2022adaptive}, like DINet~\cite{Zhang_Hu_Deng_Fan_Lv_Ding_2023}, which warp the global feature map for simplicity and efficiency. However, this can result in overly smoothed lip shapes, reducing the expressiveness of the generated videos.

\begin{figure}[t]
  \centering
  \includegraphics[width=0.7\linewidth]{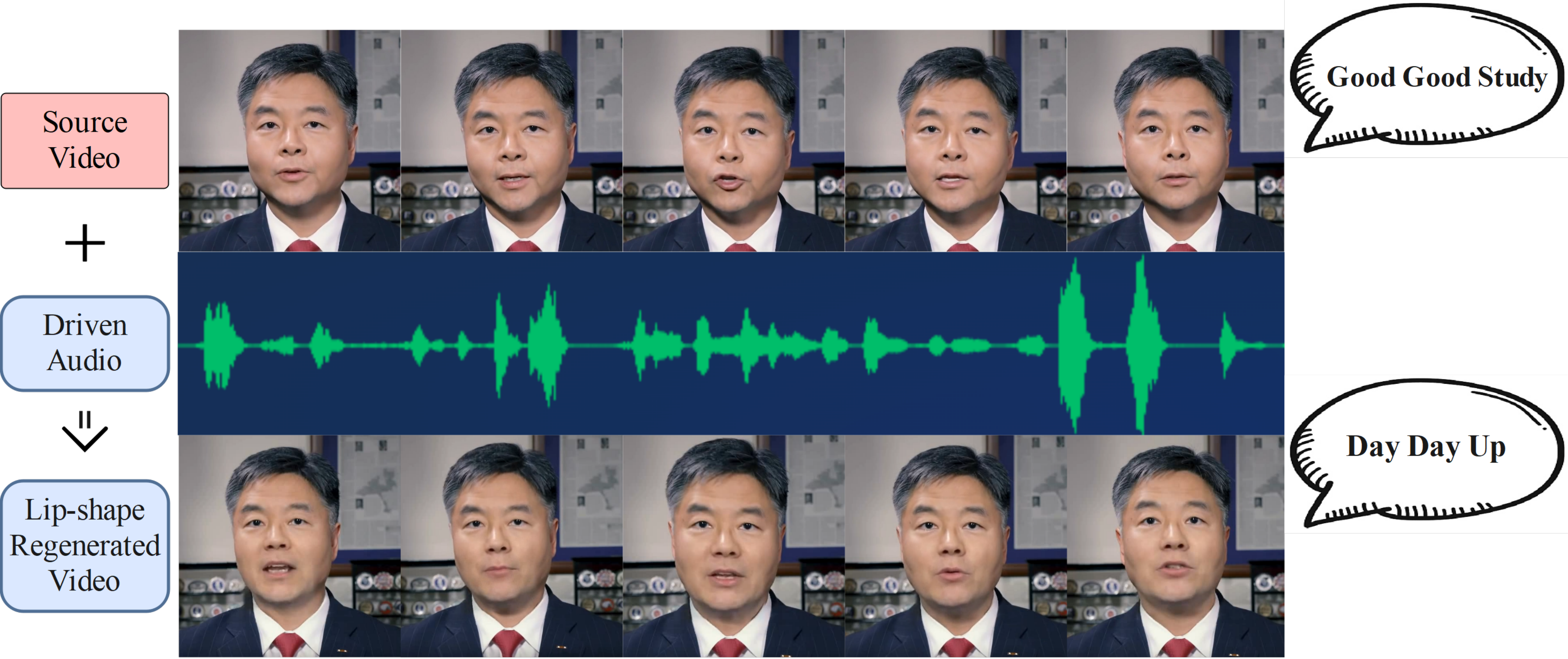}
  \vspace{-0.5 em}
  \caption{Audio-driven lip synthesis: This figure demonstrates the effect of generating new lip shapes in the source video according to the driven audio.}
  \label{fig:唇形驱动示意图}
  \vspace{-1.2 em}
\end{figure}

Despite the success of prior implicit methods, they have not fully taken into account \textit{local flexibility} in the lip area, while considering the whole areas such as the head and neck without distinction. To adaptively extract accurate lip-part features, our method uses the movement of \textit{audio-related muscles} in the lip area instead of fixed keypoints, as different muscles are engaged when different people speak different words.

\begin{figure*}[!htb]
  \centering
  \includegraphics[width=0.7\linewidth]{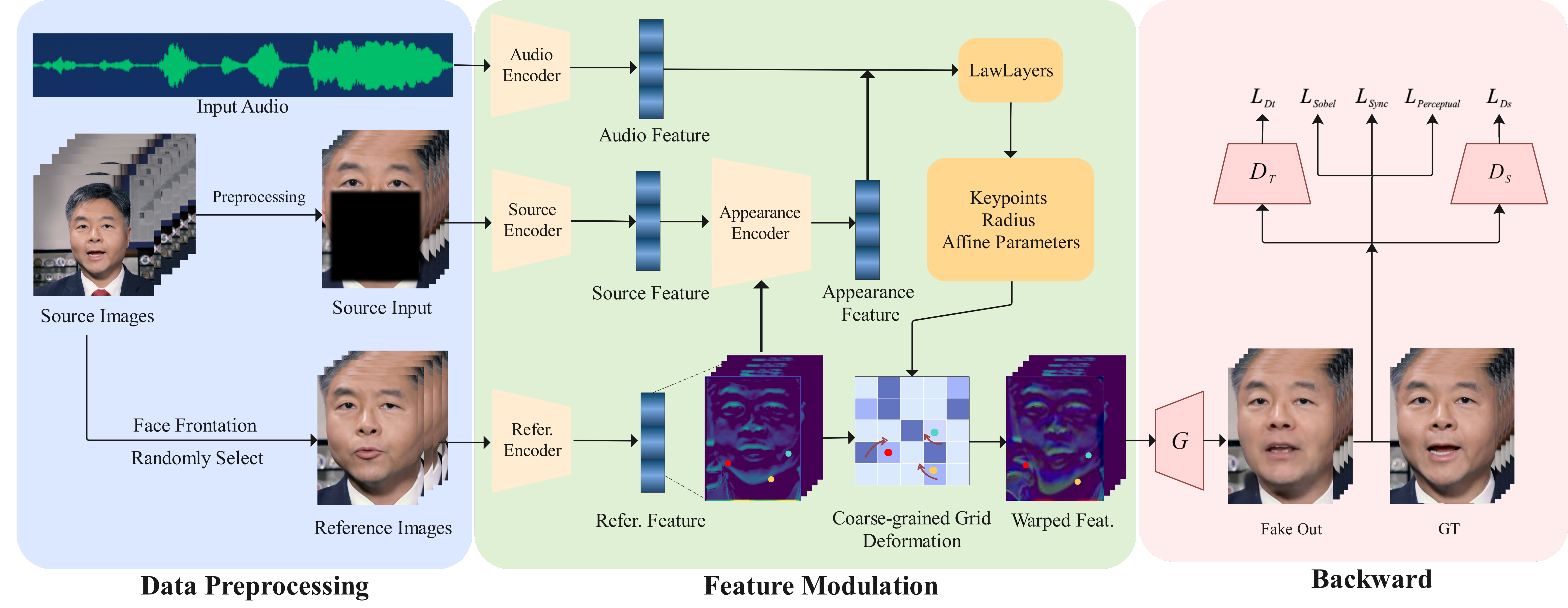}
  \vspace{-0.5 em}
\caption{Overview of the LawDNet framework: Data preprocessing aligns inputs via face frontalization and soft masking. Feature modulation, guided by audio and visual cues, uses keypoints and affine parameters for feature warping, generating lip-synced outputs via \(G\). Dual discriminators \(D_T\) and \(D_S\) and multi-level losses ensure training stability and quality.}
\label{fig:main_flowchart}
\end{figure*}

we introduce \textbf{LawDNet}, a novel architecture that provides flexibility and efficiency in modeling lip movements using local affine warping. Unlike flow-based methods~\cite{siarohin2019first,wang2021one,yin2022styleheat} that rely on first-order derivatives and dense optical flows, LawDNet balances between the benefits of learning optical flows and applying global rigid affine transformations on feature maps. It employs a set of self-learned keypoints and adaptive radii to define local affine transformations, flexibly allocating different degrees of freedom to specified lip areas. These warping fields are constructively defined using matrix-exponential operations directly on each feature map, rather than being learned by passing local affine information into a separate sub-network, as in FOMM~\cite{siarohin2019first}. This approach avoids the computational overhead associated with dense optical flows while enabling non-linear warping of feature maps. Using low-dimensional warp parameters within the coarse-grained grid enables efficient lip-sync video generation.

Previous works~\cite{DBLP:journals/corr/abs-2008-10010,wang2023seeing} use 2D discriminators, lacking temporal consistency in lip movements. We propose a dual-stream architecture with a Spatial Discriminator and a Temporal Discriminator, along with an in-the-wild dataset and and develop a preprocessing pipeline for audio-visual synchronization and facial frontalization.
Our key contributions are:
\begin{enumerate}
\item LawDNet, leveraging self-learned keypoints and local affine Warping deformation on feature map for zero-shot lip synthesis.
\item Developed a dual-stream discriminator with a temporal module for consistent lip movements.
\item Corrected a dataset with diverse scenarios and a preprocessing pipeline for synchronization and frontalization.
\end{enumerate}

\section{Method}

Audio-driven lip synthesis involves generating a coherent sequence of facial images \( \{I_t\}_{t=1}^{T} \) that correspond to a given audio input \( A \). The complete pipeline is illustrated in \cref{fig:main_flowchart}.
\vspace{-0.1 em}

\subsection{Data Preprocessing}
\label{Chapter one data preprocessing}

Our preprocessing pipeline includes \textbf{audio-visual synchronization} and \textbf{facial frontalization}. For synchronization, we use SyncNet~\cite{Chung16a} to detect and correct any misalignment between audio and video.

\begin{figure}[t]
  \vspace{-1.0 em}
  \centering
  \includegraphics[width=0.6\linewidth]{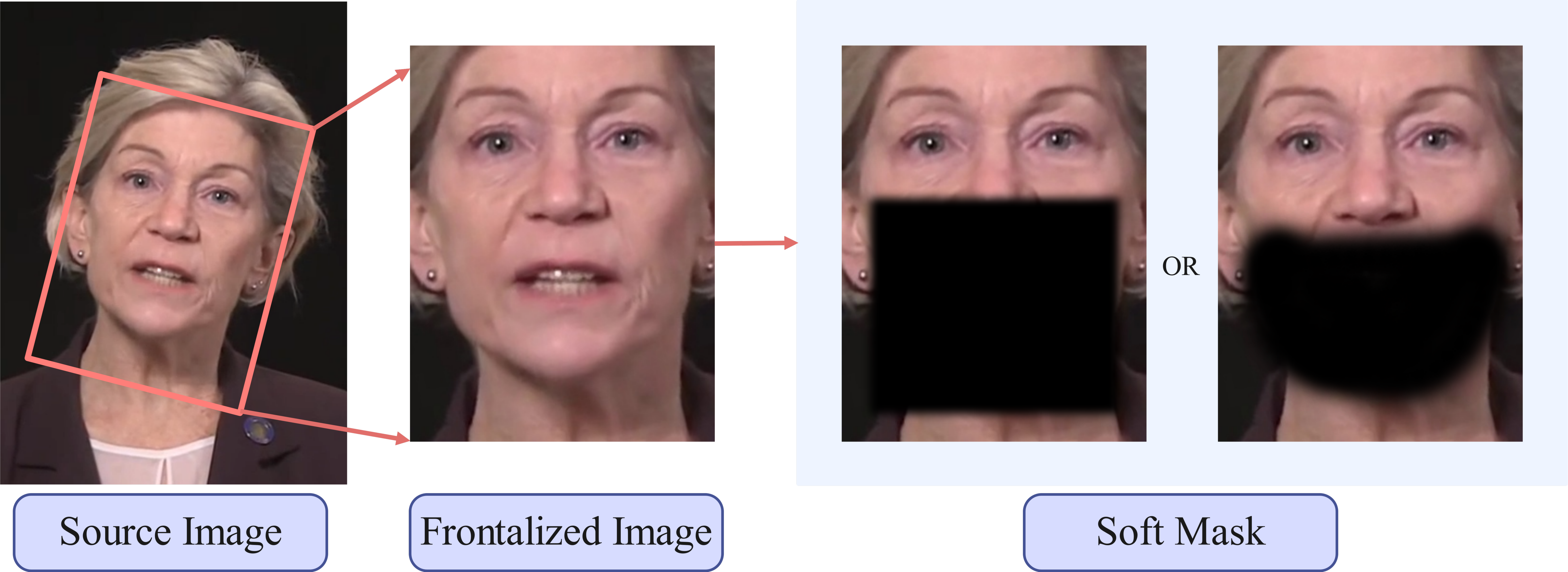}
    \vspace{-0.5 em}
   \caption{Illustration of the data preprocessing.}
  \label{fig:正脸化示意图}
  \vspace{-0.8 em}
\end{figure}

For facial frontalization, we calculate a similarity transformation matrix \( T \) to align detected facial landmarks \( M \) with a frontal landmark template \( M_{\text{tmp}} \), defined as:
\begin{equation}
T = [sR,t] = \arg\min_{s,R,t}\|s\cdot R\cdot M +t - M_{\rm tmp}\|,
\end{equation}
where \( s \), \( R \), and \( t \) represent scaling, rotation, and translation, respectively. Applying \( T \) produces the frontalized image \( I_{\text{front}} = T \cdot I \). The original head pose can be restored with the inverse transformation \( T^{-1} \). Finally, a mask with Gaussian blur is applied around the lip area to ensure smooth transitions and natural-looking lip movements, as illustrated in \cref{fig:正脸化示意图}.

\begin{figure}[!htb] 
  \vspace{-1.0 em}
  \centering
  \includegraphics[width=\columnwidth]{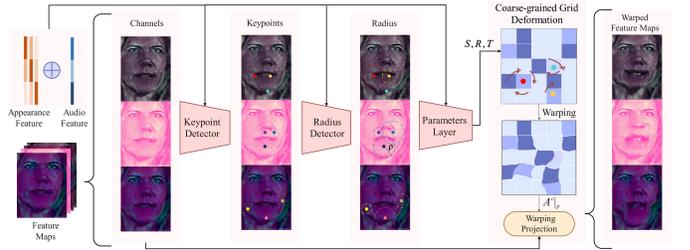} 
  
\caption{Illustrations of the local affine warping Deformation process applied to feature maps, with keypoints, radii (\(\rho\)), and affine transformation parameters (\( S,R,T\)) learned under the guidance of audio-visual features. Each module generates a distinct warping field \(A^c|_p\) on the coarse-grained grid, effectuating the deformation of the feature maps across individual channels.}
\label{fig:law}
\vspace{-1.3 em}
\end{figure}

\subsection{Local Affine Warping Deformation}
\label{Chapter 2 lawdnet}
When we take a closer look at the structure of conventional warping-based neural networks in image-driving tasks. Optical flow based methods~\cite{yin2022styleheat,siarohin2019first,zhang2023metaportrait} require the network to learn a precise registration among pixels, often causing unexpected distortion caused by warping fields over-high freedoms. Meanwhile, pixel-wised warping fields with dense representations put higher pressure on modeling and computational costs. In contrast, methods~\cite{zhang2022adaptive,Zhang_Hu_Deng_Fan_Lv_Ding_2023} that employ specific spatial transform operators often learn a global affine matrix for each discrete feature layer. These approaches are more stable and cost-free to some extent due to the efficient discrete representations of transformations acting on latent spaces. However, the low degree of freedom and the inflexible linear form limit the network's capabilities to represent complex transformations by merely combining global linear transforming on features.

What truly differentiates LawDNet from its counterparts is how it determines the globally nonlinear warping transformations of feature maps via sparse representations. We call this newly proposed technique \textbf{Local Affine Warping Deformation}. For each specific feature layer \( F^c \), \( c \) denotes the number of channels in the feature map, we first locate several abstract keypoints \( k^c_i \), $i = 1,\dots, N$, $N$ the hyper-parameter determining the number of keypoints, endowed with parameters of the local affine \( A^c_i \in SE(2)\) \cite{schneider2002geometric} operating at  \( k^c_i \). Additionally, an impacting radius factor \( \rho^c_i \) is learned to determine the extent to which \( k^c_i \) influences the transformation of points \( p \) within its neighborhood. The entire warping field can be defined as
\vspace{-0.5 em}
\begin{equation} 
    A^c|_p =  \exp( \sum_i^N e^{-\rho^c_i\|p-k^c_i\|^2}A_i^c),
\vspace{-0.1 em} 
\end{equation}
where \( \exp \) donates the exponential of matrics. In the above way, arbitrary points on the feature image will be tracked to move smoothly by the learned affine warping transformation of keypoints. One can imagine it as a learnable Liquify~\cite{wolper2017photograph} of the abstract features, as shown in~\cref{fig:law}. This is a pivotal aspect of our model, as it admits a highly controllable non-linear transformation of each point in the feature map. For computational convenience, we obtain the final warping by compositing the deformation vector field \( \Delta p^c \) with the distant-aware Softmax average, rather than computing the consuming matrix exponentials. Mathematically, 
\vspace{-0.1 em}
\begin{align}
\vspace{-1.0 em}
    \Delta p^c &= \sum_i^N\omega^c_i* (e^{-\rho^c_i\|p-k^c_i\|^2})*\Delta p^c_i,\\
    \Delta p^c_i &= p^c_i -p, \\
    p^c_i &= S^c_i(p) \cdot R^c_i(p) \cdot (p - k^c_i) + k^c_i + T^c_i(p),\\
    \omega^c_i &= {\rm Softmax}(e^{-\rho^c_i\|p-k^c_i\|^2}),
\label{eq:important}
\vspace{-0.3 em}
\end{align}

where  $S^c_i,R^c_i,T^c_i$  $k^c_i$ are the learnable affine parameters and $\rho^c_i$ determines the weights corresponding to each keypoint $k_i^c$. By applying independent local affine transformations on each channel of the feature map, LawDNet effectively models complex lip movements with greater flexibility and precision.

To reduce computational costs, we introduce a \textbf{Coarse-grained Grid Deformation} technique that minimizes memory usage while maintaining performance. Instead of direct warping on the original feature map \( F^c \), we compute the warping field on a downscaled coarse grid \( G^c \), as follows:
\begin{align}
    & G^c = \text{CoarseGrain}(F^c, \text{downscale\_factor}), \\
    & \hat{A}^c|_p =  \exp( \sum_i^N e^{-\rho^c_i\|p-k^c_i\|^2} A_i^c \text{ on } G^c),
\vspace{-0.9 em}
\end{align}
\vspace{-0.4 em}

where \( G^c \) is the downscaled grid and \( \hat{A}^c|_p \) is the affine transformation at each point \( p \). The warped grid is then upscaled to match the original feature map:
\begin{align}
\vspace{-1.0 em}
    & A^c|_p = \text{Upscale}(\hat{A}^c|_p, \text{upscale\_factor}),
\end{align}
\vspace{-0.2 em}
preserving feature map integrity without loss of detail.

\subsection{Dual Discriminators}
\label{Chapter three double flow discriminator}

GANs~\cite{goodfellow2014generative} are effective for generating lip synthesis videos~\cite{DBLP:journals/corr/abs-2008-10010,yin2022styleheat}, but often lack temporal consistency across frames. We introduce a dual-discriminator design: a Spatial Discriminator ($D_S$) with 2D convolutions to evaluate the spatial quality of individual frames, and a Temporal Discriminator ($D_T$) with 3D convolutions to capture temporal coherence across the sequence \( \{I_t\}_{t=1}^{T} \). The fake samples include random combinations of ground truth and generated video sequences, which help the model learn to differentiate between realistic and synthesized content, ensuring both spatial fidelity and smooth frame transitions.

\subsection{Loss Function}
\label{Chapter 4 Loss function}

The total loss integrates several components: Temporal Discriminator Loss \(L_{Dt}\) for frame transitions, Spatial Discriminator Loss \(L_{Ds}\) based on PatchGAN~\cite{isola2017image} for spatial coherence, Perceptual Loss \(L_{Percep}\)~\cite{johnson2016perceptual} for visual fidelity, Lip Synchronization Loss \(L_{Sync}\)~\cite{Chung16a} for lip-audio alignment, and Sobel Loss \(L_{Sobel}\) for edge clarity~\cite{996}.

\begin{figure}[!htb] 
  \centering
  \includegraphics[width=\columnwidth]{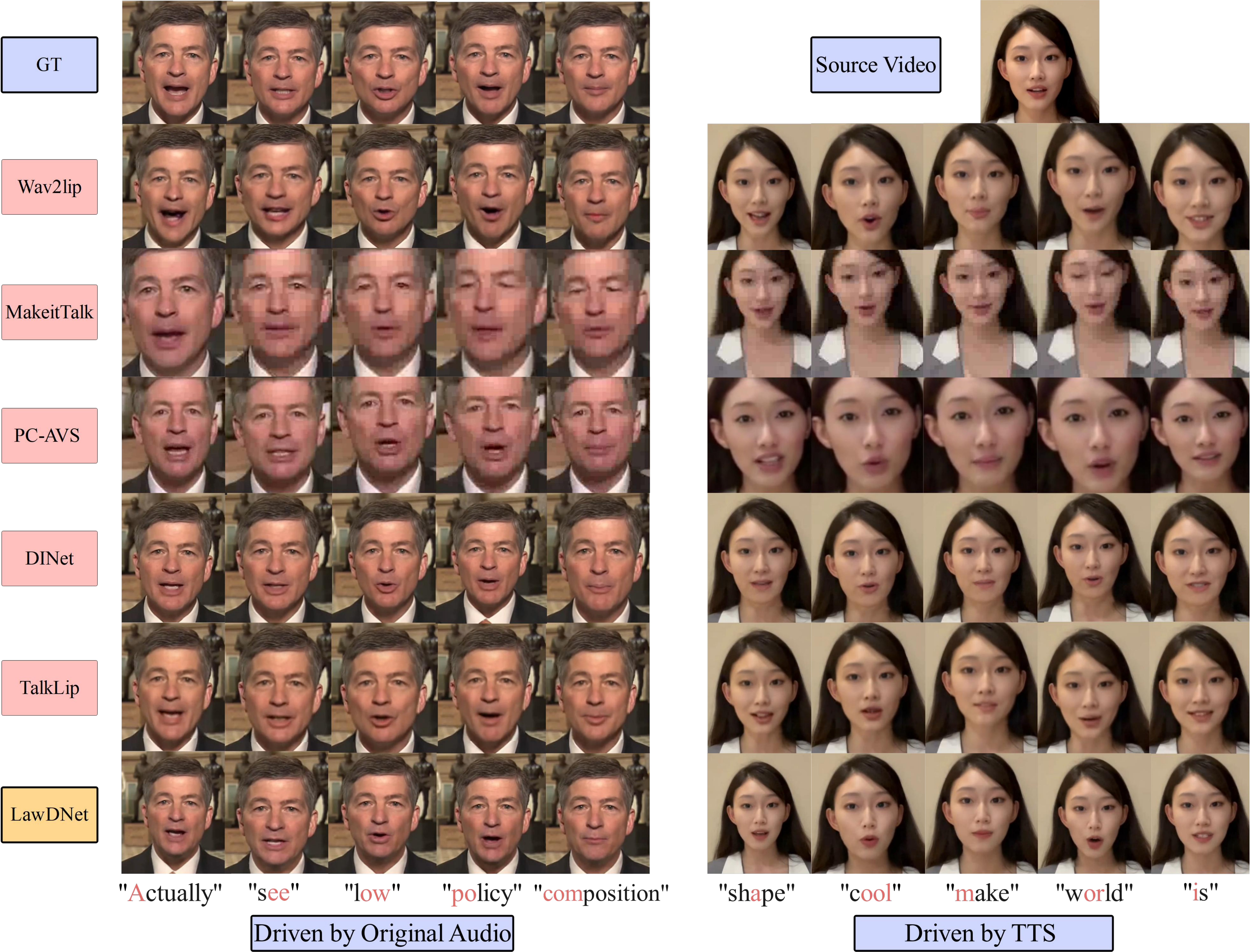} 
  \vspace{-0.8 em}
  \caption{Qualitative comparison with SOTAs. The left images are generated using the original audio from the source videos, with the goal of closely replicating the ground truth lip shapes. The right images display lip synchronization with Text-to-Speech (TTS) audio, evaluating the model's generalization capabilities. Note: 'MakeItTalk~\cite{zhou2020makelttalk}' and 'PC-AVS~\cite{zhou2021pose}' require additional head and eye motion input, hence the discrepancy in pose.}
  \label{fig:sota方法对比图}
\end{figure}
\vspace{-0.5 em}

\section{Experiments}

\subsection{Dataset and Metrics}
We corrected the \textbf{Chinese Authentic Talking Head Dataset} from Bilibili and TikTok, featuring diverse expressions, mouth movements, and languages, ideal for training lip-sync models (\cref{tab:dataset_comparison}). Additionally, we used the \textbf{HDTF Dataset}~\cite{zhang2021flow} for high-resolution lip movement analysis. LawDNet and SOTA models were trained and tested on both datasets for fair comparison, using SSIM~\cite{wang2004image}, PSNR, LPIPS~\cite{zhang2018unreasonable}, LSE-C, and LSE-D~\cite{Chung16a} for evaluation. LawDNet is trained for 12 hours on an RTX 4090 GPU with a coarse-to-fine strategy.


\begin{table}
  \centering
  \scriptsize 
  \vspace{-0.8 em}
  \caption{Overview of Audio-Visual Talking Head Datasets}
  \label{tab:dataset_comparison}
  \newcolumntype{Z}{>{\scriptsize\centering\arraybackslash}X}

  \begin{tabularx}{\columnwidth}{@{}lccZccc@{}}
    \toprule
    Dataset & Envir. & Resolution & Subjects & Hours & Language \\
    \midrule
    LRW\cite{chung2017lip} & Wild & 360P-480P & 1k+ & 173 & English \\
    Voxceleb1\cite{nagrani2017voxceleb} & Wild & 360P-720P & 1251 & 352 & Multilingual \\
    Voxceleb2\cite{chung2018voxceleb2} & Wild & 360P-720P & 6112 & 2442 & Multilingual \\
    GRID\cite{cooke2006audio} & Lab & 720x576 & 34 & 27.5 & English \\
    RAVDESS\cite{livingstone2018ryerson} & Lab & 1280x1024 & 24 & 7 & English \\
    MEAD\cite{wang2020mead} & Lab & 1920x1080 & 60 & 40 & English \\
    HDTF\cite{zhang2021flow} & Wild & 720P-1080P & 300+ & 15.8 & English \\
    CATHD(Ours) & Wild & 720P-1080P & 400+ & 40 & Multilingual \\
    \bottomrule
  \end{tabularx}

  \vspace{-1.5 em}
\end{table}

\begin{table*}[htbp]
\centering
\scriptsize 
\caption{Quantitative comparison with other state-of-the-art methods.}
\vspace{-0.8em}
    \label{tab:SOTA定量对比实验}
    
\begin{tabularx}{\textwidth}{l >{\centering\arraybackslash}X >{\centering\arraybackslash}X >{\centering\arraybackslash}X >{\centering\arraybackslash}X >{\centering\arraybackslash}X c >{\centering\arraybackslash}X >{\centering\arraybackslash}X >{\centering\arraybackslash}X >{\centering\arraybackslash}X >{\centering\arraybackslash}X}
\hline
& \multicolumn{5}{c}{HDTF} & & \multicolumn{5}{c}{Chinese Authentic Talking Head Dataset} \\
\cline{2-6} \cline{8-12}
& SSIM↑ & PSNR↑ & LPIPS↓ & LSE-D↓ & LSE-C↑ & & SSIM↑ & PSNR↑ & LPIPS↓ & LSE-D↓ & LSE-C↑ \\
\hline
Wav2Lip\cite{DBLP:journals/corr/abs-2008-10010}  & 0.9123 & 28.6549 & 0.0438 & 7.6654 & 7.0987 & & 0.9132 & 27.5688 & 0.0726 & 7.7650 & 8.0986 \\
MakeitTalk\cite{zhou2020makelttalk} & 0.8750 & 23.2816 & 0.1267 & 10.8765 & 5.1234 & & 0.8760 & 24.2355 & 0.0566 & 11.1751 & 6.1233 \\
PC-AVS\cite{zhou2021pose} & 0.9071 & 24.5563 & 0.1356 & 7.4543 & 8.0112 & & 0.9002 & 25.4577 & 0.0665 & 7.0540 & 8.2361 \\
DINet\cite{Zhang_Hu_Deng_Fan_Lv_Ding_2023}  & 0.9192 & 29.7618 & 0.0581 & 8.4432 & 8.2713 & & 0.9312 & 29.1668 & 0.0452 & 7.5430 & 6.8776 \\
TalkLip\cite{wang2023seeing}  & 0.9086 & 27.6789 & 0.0501 & 8.2328 & 7.0426 & & 0.9166 & 28.7723 & 0.0659 & 8.4320 & 7.0165 \\
\hline
GT  & 1.0000 & N/A & 0.0000 & 6.0478 & 8.9452 & & 1.0000 & N/A & 0.0000 & 5.9165 & 8.6954 \\
\textbf{LawDNet (Ours)} & \textbf{0.9375} & \textbf{30.6741} & \textbf{0.0325} & \textbf{6.9824} & \textbf{8.2921} & & \textbf{0.9417} & \textbf{30.1384} & \textbf{0.0442} & \textbf{6.8316} & \textbf{8.4603} \\
\hline
\end{tabularx}
\vspace{-0.5 em}
\end{table*}


\begin{table}[htbp]
\centering
\scriptsize
\vspace{-0.5 em}
\caption{Impact of coarse-grid size and keypoint number}
\vspace{-0.5em}
\label{tab:keypoints消融实验}
\begin{tabular}{|c|c|c|c|c|}
\hline
{Keypoint Num.} & \multicolumn{4}{c|}{Coarse-grid Size (LSE-C $\uparrow$ / LSE-D $\downarrow$)} \\ \cline{2-5}
                                  & 20x20 & 40x40 & \textbf{60x60} & Origin Size \\ \hline
\textbf{8}                        & 7.54 / 7.17 & 7.63 / 6.84 & \textbf{8.16} / \textbf{6.83} & 8.04 / 7.10 \\ \hline
5                                 & 7.49 / 7.21 & 7.60 / 7.03 & 8.10 / 7.07 & 7.82 / 7.13 \\ \hline
3                                 & 7.03 / 7.24 & 7.16 / 7.12 & 7.84 / 7.18 & 7.54 / 7.28 \\ \hline
\end{tabular}
\vspace{-1.0em}
\end{table}

\subsection{Comparison with State-of-the-Art Methods}
We compared LawDNet with zero-shot audio-driven lip synthesis methods including Wav2Lip~\cite{DBLP:journals/corr/abs-2008-10010}, MakeitTalk~\cite{zhou2020makelttalk}, PC-AVS~\cite{zhou2021pose}, DINet~\cite{Zhang_Hu_Deng_Fan_Lv_Ding_2023} and TalkLip~\cite{wang2023seeing}. As shown in~\cref{tab:SOTA定量对比实验}, LawDNet outperforms existing approaches on both the HDTF and Chinese Authentic Talking Head Dataset. On the HDTF dataset, LawDNet achieved a \textbf{3.07\%} increase in SSIM (29.7618 to 30.6741) and a \textbf{6.33\%} decrease in LSE-D (7.4543 to 6.9824). Similarly, on CATHD, it improved SSIM by \textbf{3.32\%} (29.1688 to 30.1384) and reduced LSE-D by \textbf{3.15\%} (7.0540 to 6.8316). Qualitative results in~\cref{fig:sota方法对比图} demonstrate LawDNet's superior lip-sync accuracy and liveliness. Our method excels with both original and synthesized audio, including text-to-speech (TTS) outputs, highlighting its robust generalization and applicability in diverse high-fidelity lip-sync scenarios.





\subsection{Ablation Study}

\textbf{Feature Map Warping Methods.} Our method reduces computational complexity compared to optical flow by having fewer degrees of freedom and lower parameter requirements,
\vspace{-0.4 em}
\begin{equation}
    D_{\rm global\ affine} < k \times D_{\rm global\ affine} \approx D_{\rm ours} \ll D_{\rm optical\ flow},
    \label{eq:my_equation}
\end{equation}

\noindent where \(D\) is the degree of freedom and \(k\) the number of keypoints. Keypoints require 2 degrees of freedom, while radius, scaling (S), rotation (R), and translation (T) require 1, 2, 1, and 2, respectively. This simplified parameterization eases learning compared to optical flow. Local Affine warping outperforms Global Affine and Optical Flow in LSE-D and LSE-C metrics, offering better temporal coherence and natural facial animation at relatively lower computational costs (Table~\ref{tab:warping_comparison}).

\vspace{-0.3 em}
\begin{table}[h]
\vspace{-0.4em}
\centering
\caption{Complexity and Performance Comparison of Warping Methods}
\vspace{-0.5em} 
\label{tab:warping_comparison}
\scriptsize
\begin{tabular}{@{}ccccc@{}}
\toprule
Warping Method & Param. No. & FPS & LSE-D↓ & LSE-C↑ \\ \midrule
Global Affine & \textbf{1.25 MB} & \textbf{31} & 8.1781\% & 8.2495\% \\
Local Affine & 3.75 MB & 26 & \textbf{6.9824}\% & \textbf{8.2921}\% \\
Optical Flow & 14.9 MB & 12 & 7.0736\% & 7.9308\% \\ \bottomrule
\end{tabular}
\end{table}

\begin{figure}[t]
  \centering
  \includegraphics[width=1.0\linewidth]{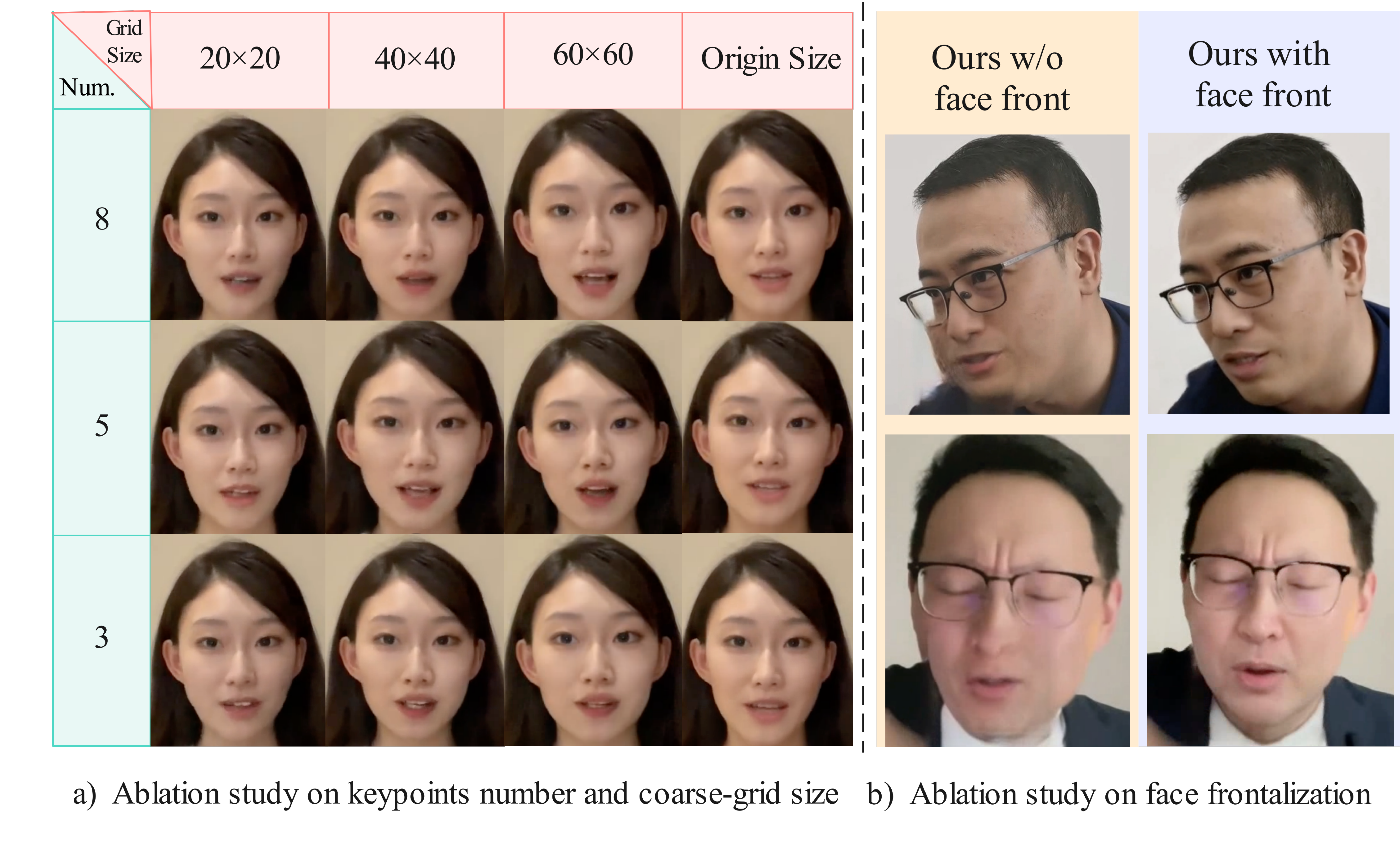}
  \vspace{-1.0 em}
    \caption{Ablation studies: (a) Coarse-Grid Size and Keypoint Numbers; (b) Face Frontalization}
  \label{fig:ablation_study face frontalization and keypoint}
  \vspace{-1.0 em} 
\end{figure}

\textbf{Grid Size and Keypoint Numbers}. Grid size impacts video quality: smaller grids enhance sharpness but can cause distortions, while larger grids improve fidelity but reduce sharpness. Fewer keypoints restrict lip movement. Our tests (\cref{tab:keypoints消融实验}) found that a grid size of \textbf{60} and \textbf{8} keypoints provide the best balance for expressive and stable videos (\cref{fig:ablation_study face frontalization and keypoint}-a).

\begin{table}[h]
\vspace{-1.0 em}
\centering
\caption{Impact of Temporal Coherence Discriminator(TCD) and Face Frontalization(FF)}
\vspace{-0.8em}
\label{tab:face_frontalization}
\scriptsize
{%
\begin{tabular}{@{}lccccc@{}}
\toprule
Method & SSIM↑ & PSNR↑ & LPIPS↓ & LSE-D↓ & LSE-C↑ \\ \midrule
TCD + FF & \textbf{0.9375} & \textbf{30.6741} & \textbf{0.0325} & \textbf{6.9824} & \textbf{8.2921} \\
TCD Only & 0.9024 & 28.9626 & 0.0414 & 8.1432 & 7.9521 \\
FF Only & 0.8941 & 28.6573 & 0.0480 & 7.5112 & 8.0187 \\
None & 0.8315 & 27.1128 & 0.0624 & 8.9842 & 7.6421 \\ \bottomrule
\end{tabular}%
}
\vspace{-0.5 em}
\end{table}

\begin{figure}[t]
  \centering
  \includegraphics[width=0.7\linewidth]{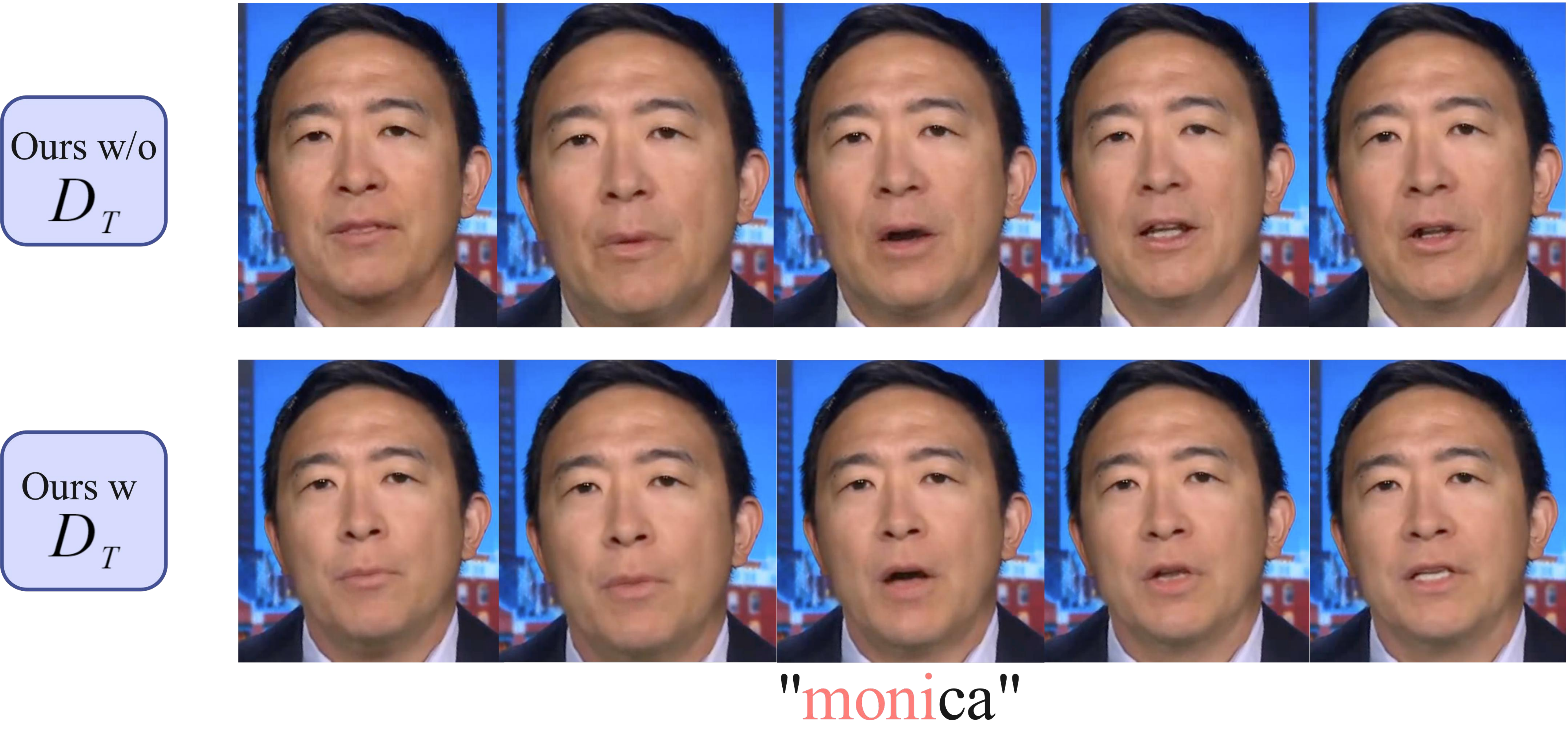}
  \vspace{-1.0 em}
  \caption{Ablation study on $D_T$}
  \label{fig:ablation_study on Dt}
\end{figure}

\textbf{Temporal Coherence Discriminator}. As shown in \cref{fig:ablation_study on Dt} and \cref{tab:face_frontalization}, using only the spatial discriminator $D_S$ yields high lip clarity but causes jittering. Adding the Temporal Coherence Discriminator (TCD) improves frame continuity and LSE-D and LSE-C scores without sacrificing image quality.

\textbf{Face Frontalization}. As shown in \cref{fig:ablation_study face frontalization and keypoint}-b, face frontalization enhances lip shape naturalness, particularly in extreme poses like side views. Without it, lip shapes often distort due to limited training samples. Table \ref{tab:face_frontalization} shows that incorporating frontalization improves all metrics, proving its effectiveness.

 \vspace{-1.0 em}

\section{Conclusion}
LawDNet enhances the vividness and temporal continuity of audio-driven lip synthesis by using a novel local affine warping approach. However, edge distortions in masked areas still pose challenges. Future work includes integrating audio-to-3D model conversion techniques~\cite{10.1145/3581783.3611734,xing2023codetalker} for better lip-reading accuracy and exploring LawDNet’s spatial handling in motion transfer and facial reenactment.



\bibliographystyle{IEEEtran}
\bibliography{references}

\end{document}